\pdfoutput=1
\documentclass{article} %
\usepackage{iclr2027_conference,times}

\usepackage{amsmath,amsfonts,bm}

\def\eqref#1{equation~\ref{#1}}

\def\1{\bm{1}}

\DeclareMathAlphabet{\mathsfit}{\encodingdefault}{\sfdefault}{m}{sl}
\SetMathAlphabet{\mathsfit}{bold}{\encodingdefault}{\sfdefault}{bx}{n}

\usepackage{hyperref}
\usepackage{url}
\usepackage{xcolor}
\usepackage{colortbl}
\definecolor{rankone}{HTML}{CFB0F9}
\definecolor{ranktwo}{HTML}{E5D3FC}
\definecolor{rankthree}{HTML}{F5EFFE}
\definecolor{linkblue}{rgb}{0,0.35,0.7}
\definecolor{baselineblue}{HTML}{1E66F5}
\definecolor{ourspurple}{HTML}{8839EF}
\definecolor{boxlavender}{HTML}{F4F5FF}
\definecolor{framelavender}{HTML}{7287FD}
\definecolor{boxgreen}{HTML}{F2F8F0}
\definecolor{framegreen}{HTML}{40A02B}
\definecolor{boxyellow}{HTML}{FDF7EF}
\definecolor{frameyellow}{HTML}{DF8E1D}
\definecolor{boxflamingo}{HTML}{FCF3F3}
\definecolor{frameflamingo}{HTML}{DD7878}
\usepackage{tcolorbox}
\newtcolorbox{answerbox}[2]{colback=#1, colframe=#2!45!white, boxrule=0.6pt, arc=5pt, left=6pt, right=6pt, top=3pt, bottom=3pt, before skip=8pt, after skip=8pt}
\hypersetup{colorlinks=true, citecolor=linkblue, linkcolor=linkblue, urlcolor=linkblue}
\usepackage{graphicx}
\usepackage{wrapfig}
\usepackage{capt-of}
\usepackage{booktabs}
\usepackage{amssymb}
\usepackage{multirow}
\usepackage[nameinlink]{cleveref}
\crefname{figure}{Fig.}{Figs.}
\Crefname{figure}{Figure}{Figures}
\crefname{table}{Tab.}{Tabs.}
\Crefname{table}{Table}{Tables}
\crefname{section}{Sec.}{Secs.}
\crefname{subsection}{Sec.}{Secs.}
\crefname{subsubsection}{Sec.}{Secs.}
\crefname{appendix}{Appendix}{Appendices}
\crefname{subappendix}{Appendix}{Appendices}

\title{StopWatch: \\ Goal Grounding for Object-goal Navigation}

\author{%
\hspace*{-\tabcolsep}%
\begin{tabular}[t]{@{}l@{}}
Xinhao Liu\textsuperscript{1,2*}
\quad
Yunsheng Tian\textsuperscript{1\textdaggerdbl}
\quad
Yuheng Qiu\textsuperscript{1}
\\[0.4em]
Pieter Abbeel\textsuperscript{1,3\textdagger}
\quad
Xi Chen\textsuperscript{1\textdagger}
\quad
Guanya Shi\textsuperscript{1,4\textdagger}
\quad
Chen Feng\textsuperscript{1,2\textdagger}
\\[0.55em]
{\normalfont\mdseries\small
\textsuperscript{1}Amazon FAR
\quad
\textsuperscript{2}New York University
\quad
\textsuperscript{3}UC Berkeley
\quad
\textsuperscript{4}Carnegie Mellon University}
\\[0.2em]
{\normalfont\mdseries\small
\textsuperscript{*}Work done during internship at Amazon FAR
\quad
\textsuperscript{\textdagger}Equal advising
\quad
\textsuperscript{\textdaggerdbl}Project lead}
\\[0.2em]
{\normalfont\mdseries\small \url{https://stopwatch-nav.github.io}}
\end{tabular}%
}

\iclrfinalcopy %
\begin{document}

\maketitle
\lhead{Preprint}

\noindent\begin{minipage}{\textwidth}
    \centering
    \vspace{-2mm}
    \includegraphics[width=\textwidth]{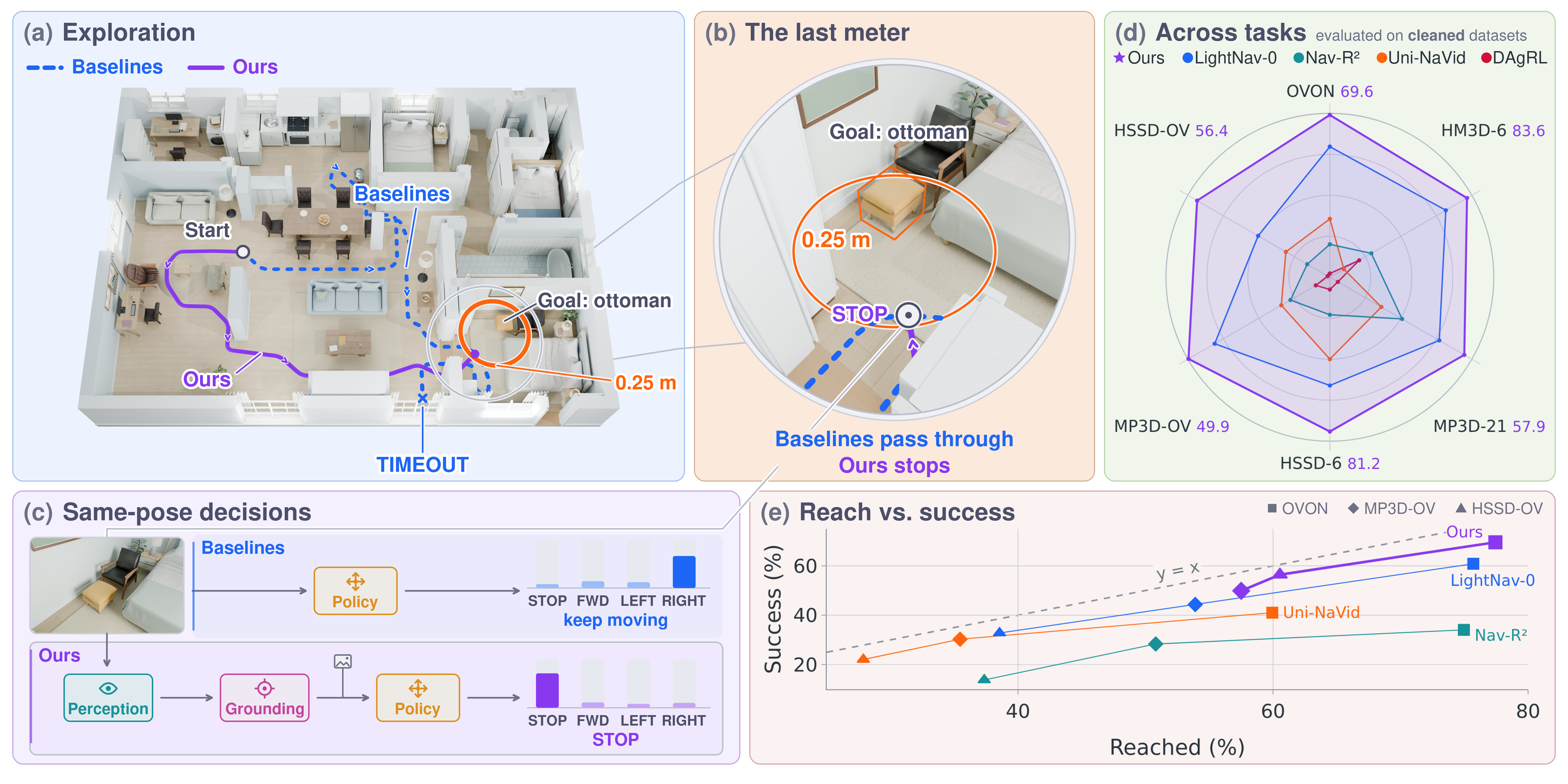}
    \vspace{-4mm}
    \captionof{figure}{
    \textbf{Goal grounding is key to object-goal navigation.}
    \textbf{(a, b)} \textcolor{baselineblue}{\textbf{Baselines}} explore as well as \textcolor{ourspurple}{\textbf{ours}}
    yet walk past the goal; ours stops within 0.25\,m.
    \textbf{(c)} Our system has explicit goal grounding, which enables the agent
    to stop at the right place.
    \textbf{(d)} Ours has the highest success rate of all compared systems on six \textit{cleaned}
    benchmarks.
    \textbf{(e)} Success rate against the rate of reaching the goal vicinity.
    All baselines reach the vicinity far more often than they succeed.
    }
    \label{fig:teaser}
\end{minipage}
\vspace{8mm}

\begin{abstract}
Pretrained VLMs bring rich world knowledge and semantic priors to object-goal navigation, and fine-tuning a VLM into a navigation policy has become a common recipe. Despite this progress, success rates remain low, and it is unclear which capability these fine-tuned policies lack. Our diagnosis identifies goal grounding as the dominant failure. These policies explore well, yet often fail to notice that the goal is already in view or to estimate how far away it is. To address this failure, we ground the goal explicitly. Explicit grounding substantially improves success and reaches state-of-the-art results on six benchmarks across three scene collections. We also correct annotation defects in these benchmarks, so that future methods are evaluated more faithfully. Our analysis on the clean benchmarks shows that, with ground-truth grounding, our policy nearly matches the expert that generated its demonstrations, leaving the headroom in perception rather than policy learning. Our work points to goal grounding as a missing piece of object-goal navigation and offers a simple and effective solution to the task.
\end{abstract}

\section{Introduction}
\label{sec:intro}

Pretrained vision-language models (VLMs) bring web-scale visual and semantic knowledge to embodied navigation.
In object-goal navigation, where an agent must find an instance of a named object category in a novel scene, fine-tuning them on expert demonstrations has become a common recipe~\citep{zhang2024uni,xiang2025navr2,chu2026abot,zhang2026qwen,wang2026lightnav,zhang2026embodied}. Progress has come from larger demonstration sets~\citep{zhang2024uni,zhang2026embodied,zhang2026qwen,chu2026abot} and better modeling of the observation history~\citep{zhang2024navid,zhang2024uni,zhang2026qwen}, yet these policies still generalize poorly to unseen goal categories. Further scaling brings diminishing returns~\citep{zhang2024uni,wang2026lightnav}, and some observe that perception and grounding, rather than the navigation policy, limit performance~\citep{xue2026omninav,chu2026abot}. What is missing is a diagnosis of which capability these policies lack. In this paper we give that diagnosis and build an object-goal navigation system around it.

\begin{wrapfigure}[23]{r}{0.3\textwidth}
    \vspace{5mm}
    \centering
    \includegraphics[width=0.3\textwidth]{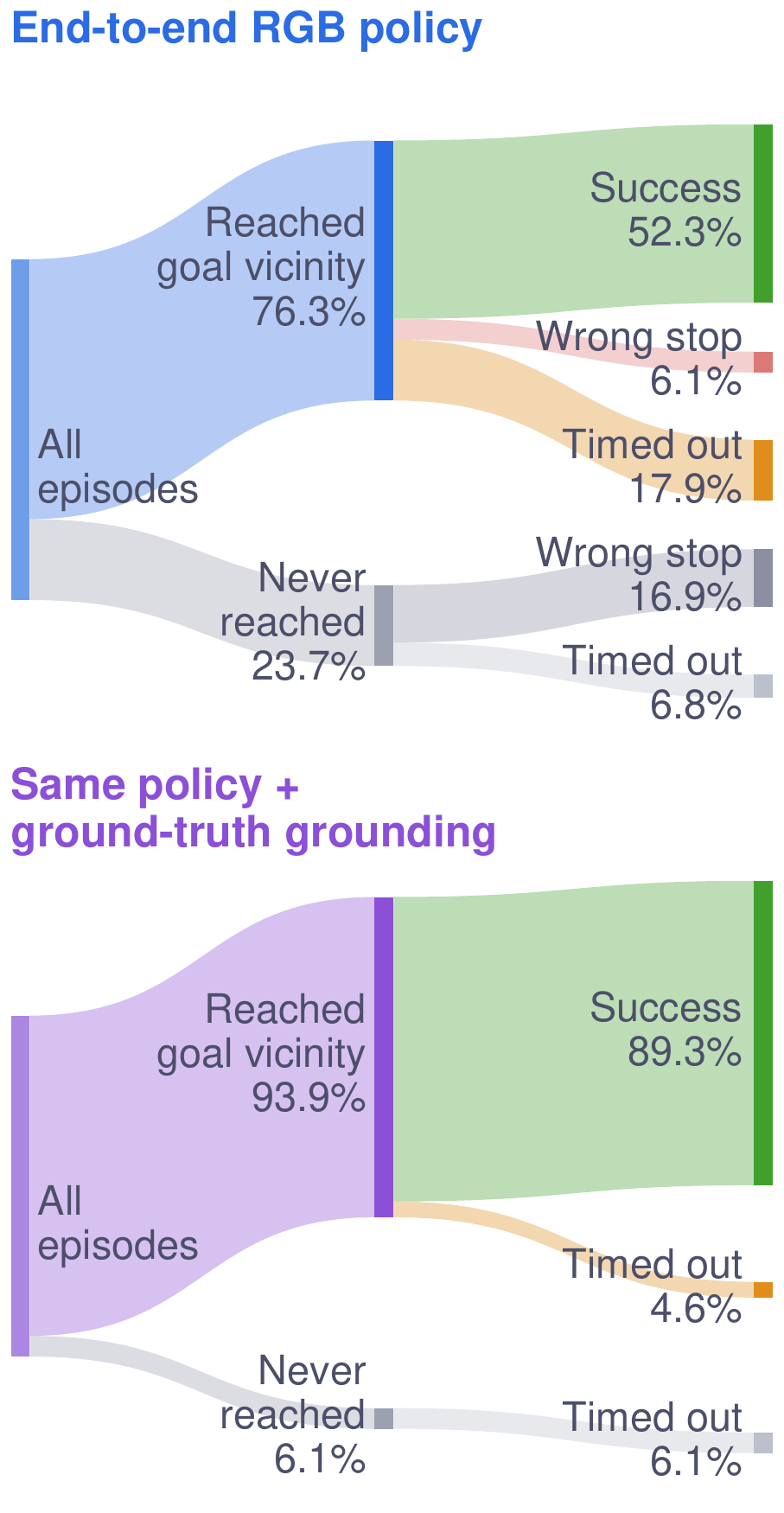}
    \vspace{-6mm}
    \caption{
    \textbf{Where episodes end.}
    We trace novel-goal episodes of an end-to-end RGB policy (top) and the same policy with ground-truth grounding (bottom).
    }
    \label{fig:failure}
    \vspace{-12mm}
\end{wrapfigure}

Our diagnosis indicates these policies fail at \textbf{goal grounding}: determining whether the goal is in view and how far it is. These policies learn search well from demonstrations. They actually reach the goal far more often than they succeed (\cref{fig:teaser}e). An end-to-end policy can come within 0.25\,m of the goal in 76\% of episodes yet succeed in \textit{only} 52\% (\cref{fig:failure}). In contrast, given the ground-truth semantics in view and distance to the goal, the same policy can have a success rate of 89\%. Stopping at the right place thus turns out to be harder than reaching it.

The diagnosis suggests a direct remedy. We make goal grounding explicit, so that before each action the agent states which objects are in view and how far the goal is, much like chain-of-thought reasoning in language models and embodied agents~\citep{wei2022chain,zawalski2024robotic,lin2025navcot}. Unlike free-form reasoning traces, which are generated by large VLMs and prone to noise~\citep{zawalski2024robotic,xiang2025navr2,wang2025auxthink,gao2025octonav}, our goal grounding is derived from the semantic and geometric labels of the training scenes. Every demonstration frame comes with its own ground-truth goal grounding at no extra labeling cost.

At test time, the grounding text must be predicted, so we build a pipeline of two models with this text as their interface. A perception model predicts the grounding from RGB observations, and a policy model reads it alongside the observations to choose the next action. The two are trained separately, the policy on ground-truth grounding and the perception model on the same annotations. Goal grounding could also be learned inside a single end-to-end model as an additional output, but such a model navigates worse in our experiments. The pipeline reaches state-of-the-art success on six object-goal navigation benchmarks, 8.8 points above the strongest baseline on HM3D-OVON val\_unseen with corrected annotations (\cref{tab:main}). These results reduce a long-standing failure of object-goal navigation to a concrete perception problem.

Our contributions are threefold:
\begin{itemize}
\item We identify \textit{goal grounding} as a missing piece for object-goal navigation. Policies often reach the goal but fail to stop, which can be mostly fixed by correct grounding.
\item Guided by this, we build a two-model pipeline that makes goal grounding explicit, reaching state-of-the-art results on all benchmarked object-goal navigation tasks.
\item We guide future work from two perspectives. We remedy annotation defects in existing benchmarks, so future methods are evaluated more faithfully. On the clean benchmarks, we show that the policy is already strong and the remaining headroom lies in perception.
\end{itemize}

\section{Related Work}

\paragraph{Embodied navigation tasks.}
Navigation tasks differ in how they specify the goal.
Early tasks give the goal as coordinates or an image~\citep{anderson2018evaluation, liu2025citywalker, zhu2017target, krantz2023navigating, liu2026wanderland}.
These tasks are less complex and addressed by large-scale reinforcement learning~\citep{wijmans2020ddppo}.
Object-goal navigation names an object category and asks the agent to find any instance of it~\citep{batra2020objectnav, yokoyama2024hm3dovon}. The agent must search the scene efficiently, remember where it has been over long horizons, and recognize the goal once it comes into view. In the open-vocabulary setting, test goals include categories never used as goals in training.
Like most recent work, we study navigation in continuous environments~\citep{zhang2024navid, zhang2024uni, cheng2024navila}, and focus on open-vocabulary object-goal navigation.

\paragraph{Modular navigation systems.}
Modular systems decompose navigation into perception and planning modules~\citep{chaplot2020semexp, ramakrishnan2022poni}.
Each module can be inspected on its own, and such systems transfer to real homes far better than end-to-end policies~\citep{gervet2023navigating}.
Open-vocabulary models~\citep{radford2021clip, liu2024groundingdino} made this line zero-shot.
Detectors localize the goal, and image-text similarity or LLM commonsense scores the frontiers to explore~\citep{gadre2023cow, zhou2023esc, yokoyama2024vlfm}.
LLMs also plan in language, parsing instructions into landmarks or reading captions of the observations~\citep{shah2023lmnav, zhou2024navgpt}.
Similarly, we make perception explicit, but only for the two cues our diagnosis finds missing.
Unlike their planners, which act only on this interface by fixed rules, our policy sees every frame and learns search and stopping from demonstrations.

\paragraph{VLMs for navigation.}
Rather than composing modules, recent work fine-tunes a single pretrained VLM into the navigation policy, relying on its world knowledge for what these tasks demand. This is now the dominant approach~\citep{zhang2024navid, cheng2024navila, zhang2024uni, xue2026omninav, chu2026abot, zhang2026qwen, wang2026lightnav, zhang2026embodied}.
These policies follow the vision-language-action (VLA) model paradigm~\citep{brohan2023rt, kim2024openvla}, where the model is trained end-to-end by behavior cloning on expert demonstrations.
However, such training has been shown to harm the model's semantic priors and language understanding~\citep{fang2025intention, driess2025knowledge, zhou2025chatvla}.
We follow this line of work, but supervise perception explicitly with structured text instead of relying on action labels alone.

\paragraph{Intermediate reasoning.}
Writing out intermediate steps before the output improves both language models~\citep{wei2022chain} and VLA models~\citep{zawalski2024robotic}.
In navigation, policies likewise generate reasoning text before acting, either at inference or as training-time supervision~\citep{lin2025navcot, gao2025octonav, wang2025auxthink}.
The content of this text is chosen by design, typically plans, subtasks, or instruction progress written by a larger model, without evidence of what the policy lacks.
Our grounding text instead holds only the two cues our diagnosis finds missing, the semantics in view and the distance to the goal, both read from the scene annotations of every training frame.

\section{StopWatch}
\label{sec:method}

A policy fine-tuned from a VLM should be well equipped for object-goal navigation. The base model supplies general visual knowledge, and millions of expert demonstrations supply the policy. Yet success on novel goals stays low, and further scaling returns little. Counterintuitively, the agent explores an unfamiliar scene like an expert, nears the goal, and does not stop. The pretrained knowledge does not turn into the right decisions. This section locates the root cause, derives our system from it, and fixes the evaluation.

\subsection{Problem setup}
\label{sec:setup}

We study object-goal navigation \citep{zhu2017target,batra2020objectnav}. In the task, an agent is dropped into an unfamiliar environment. It must find an instance of a goal category and stop near one within a step budget.
In the open-vocabulary form of the task \citep{yokoyama2024hm3dovon}, the goal at test time may be unseen in training. The task couples open-ended semantics with embodied spatial search. A household robot faces this problem when asked to fetch a towel in a home it has never seen. Following the dominant approach \citep{zhang2024navid,cheng2024navila,zhang2024uni}, we train the policy by behavior cloning (BC) on demonstrations from a frontier-exploration expert~\citep{yamauchi1997frontier}. We fine-tune a pretrained VLM to predict the expert's next action from the goal and the history of observations.

\subsection{Failure analysis of existing methods}
\label{sec:failure}

\paragraph{Reaching without stopping.}
\looseness=-1
Success in object-goal navigation has two requirements. The agent must reach the goal, and it must \texttt{STOP} there. Fine-tuned VLM policies mostly meet the first requirement and miss the second, which has long been noted as a distinct difficulty~\citep{wasserman2022lastmile,song2023terminate}. On novel goal categories, an end-to-end policy comes within 0.25\,m of a goal in 76\% of episodes, yet succeeds in only 52\% (\cref{fig:failure}). Near the goal it walks past without stopping. Most failures come from the model not knowing that it has arrived.

\paragraph{Why stopping fails.}
The two requirements pose very different problems for behavior cloning. The expert explores each scene in the same way until the goal comes into view, so the cloned search works just as well for a new category. Stopping, in contrast, requires recognizing the goal, and has much sparser supervision. A demonstration has a single \texttt{STOP} at its end, and only for training goal categories. From this one action per episode, the policy must learn goal grounding, namely whether the goal is in view and how far it is. Imitation alone does not make the policy use the VLM's pretrained knowledge, so it fails to stop for unseen goals.

\paragraph{Grounding restores stopping.}
To test whether goal grounding is what the policy lacks, we give the same policy the ground-truth grounding text before each action. On the same episodes, success rises from 52\% to 89\%, and failures after reaching the goal fall from 24\% to 5\% (\cref{fig:failure}, bottom). Wrong stops disappear entirely, and every remaining failure is a timeout. The end-to-end policy therefore fails at goal grounding. Our design must predict the grounding at test time (\cref{sec:design}).

\subsection{Design choices}
\label{sec:design}

\begin{figure}[t]
    \centering
    \includegraphics[width=\textwidth]{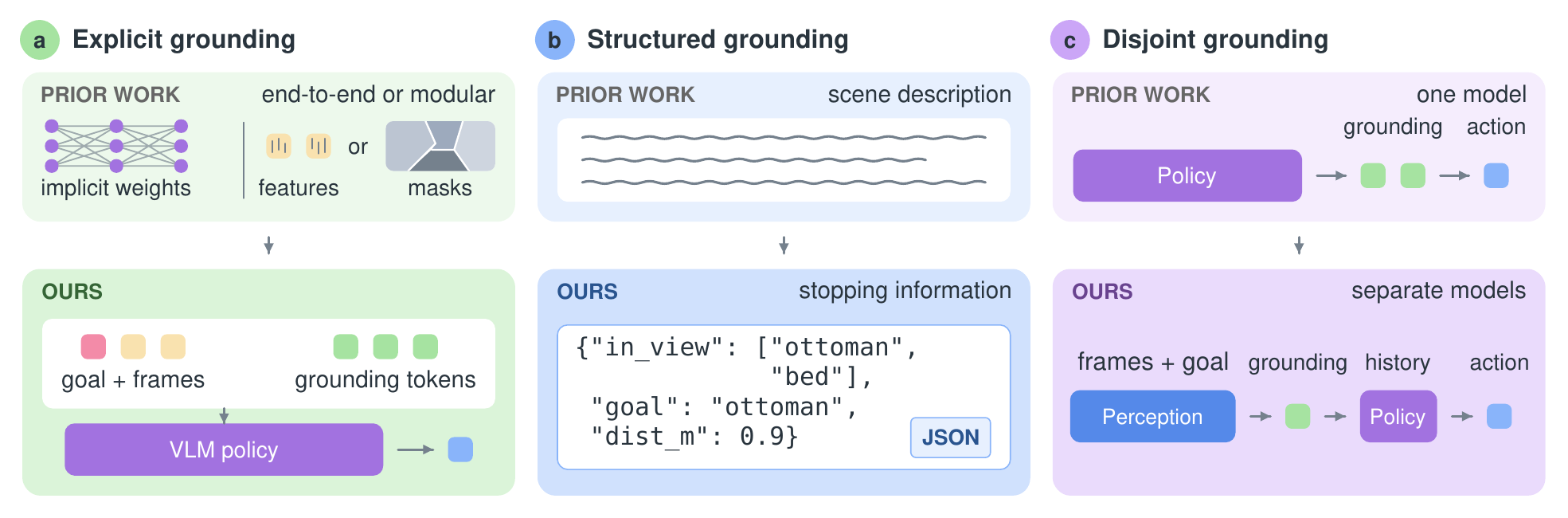}
    \caption[Design choices.]{
    \textbf{Design choices.}
    Three decisions turn an end-to-end policy into our system. Each column
    contrasts a design used in prior work (top) with our choice (bottom).
    }
    \label{fig:design}
\end{figure}

\paragraph{Explicit grounding.}
A BC-trained policy grounds the goal implicitly in its weights, and \cref{sec:failure} shows that this implicit grounding fails on novel categories. Our first choice is therefore to make the grounding explicit, written out as plain text that the policy reads before acting (\cref{fig:design}a). The policy then receives the grounding as an input at every step, instead of inferring it from the single \texttt{STOP} of each demonstration. An explicit grounding can also be inspected, which shows whether a failure comes from grounding or from acting. Modular methods carry such explicit signals as semantic maps, fused features, or segmentation masks~\citep{chaplot2020semexp,gervet2023navigating}. We are instead inspired by reasoning in VLMs~\citep{wei2022chain,zawalski2024robotic}, and carry the signal as text, the input a VLM already reads, so the policy needs no new encoder.

\paragraph{Structured grounding.}
Free-form reasoning data costs extensive labeling and quality control. Instead, we take advantage of the semantic and geometric labels from the scenes themselves. The labels compile into a structured grounding text with two fields, the object categories in view and the distance to the goal (\cref{fig:design}b). These are the two quantities that stopping needs (\cref{sec:failure}), and their ground-truth values close most of the failures after reaching the goal (\cref{fig:failure}). Because the text is compiled from annotations, every frame of every demonstration carries it at no labeling cost, so goal grounding is supervised at every frame rather than once per episode. The category list also names objects other than the goal, so its supervision covers categories that never serve as training goals.

\paragraph{Disjoint grounding.}
At test time no annotation exists, so the grounding text must be predicted. A single model could narrate and act in the chain-of-thought manner. However, we find that such a model navigates worse (\cref{tab:ablation}), so we split the system into a perception model and a policy model (\cref{fig:design}c). 
The perception model predicts the grounding text from recent frames and the goal word. The policy acts on the observations together with the grounding text, and is trained on annotated groundings. Each model thus learns one task from its own supervision. 
Perception becomes a standalone prediction problem, trained and scaled independently of the policy, and more accurate grounding text improves the system without retraining the policy (\cref{sec:analysis}).
Open-vocabulary detectors~\citep{carion2026sam3, minderer2023owlv2, liu2024groundingdino} could predict the semantic field without training, but their label sets differ from open-vocabulary annotations. In our study their less accurate predictions confused the policy.

\begin{figure}[t]
    \centering
    \includegraphics[width=\textwidth]{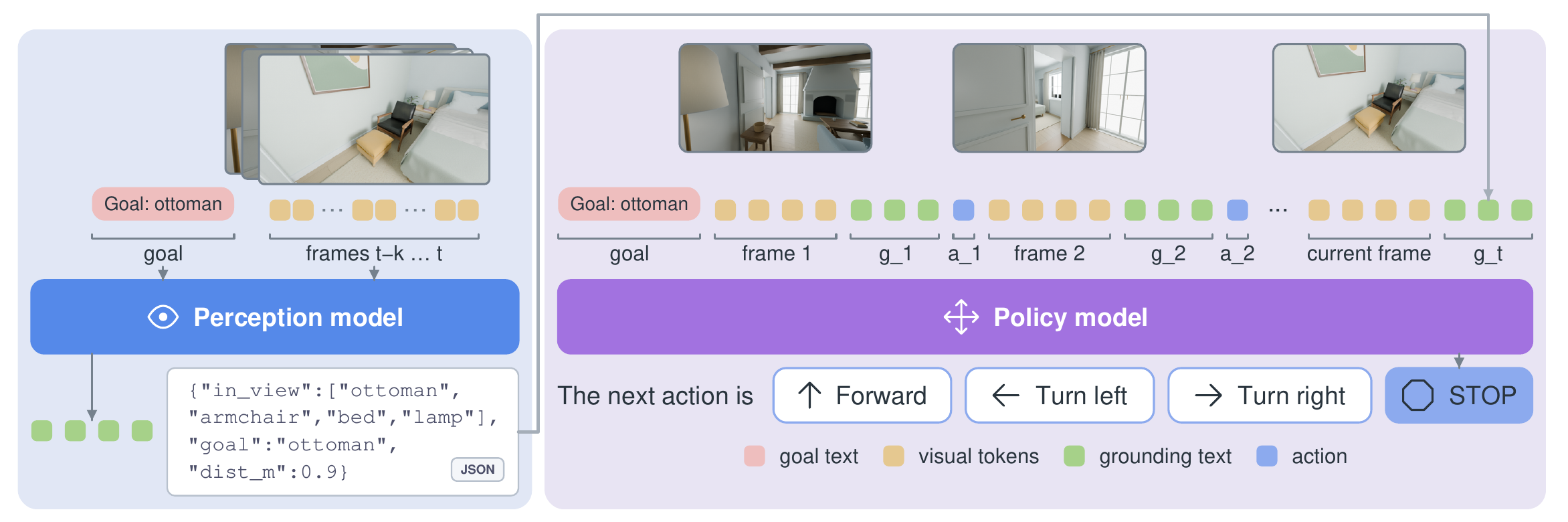}
    \caption{
    \textbf{System overview.}
    The perception model writes grounding text for each frame. The policy
    reads the frames interleaved with this text and outputs the next action.
    The two models are trained separately, both on grounding text from the
    scene annotations.
    }
    \label{fig:method}
\end{figure}

\paragraph{The system.}
\Cref{fig:method} shows the assembled system. The perception model turns each observation into grounding text. The policy reads the interleaved history of frames and grounding text and outputs the next action. Swapping the predicted text for ground truth gives the upper bound of the design, and the gap between the two runs measures the cost of imperfect perception.

\subsection{Unreliable evaluation}
\label{sec:vocab}

\begin{figure}[t]
    \centering
    \includegraphics[width=\textwidth]{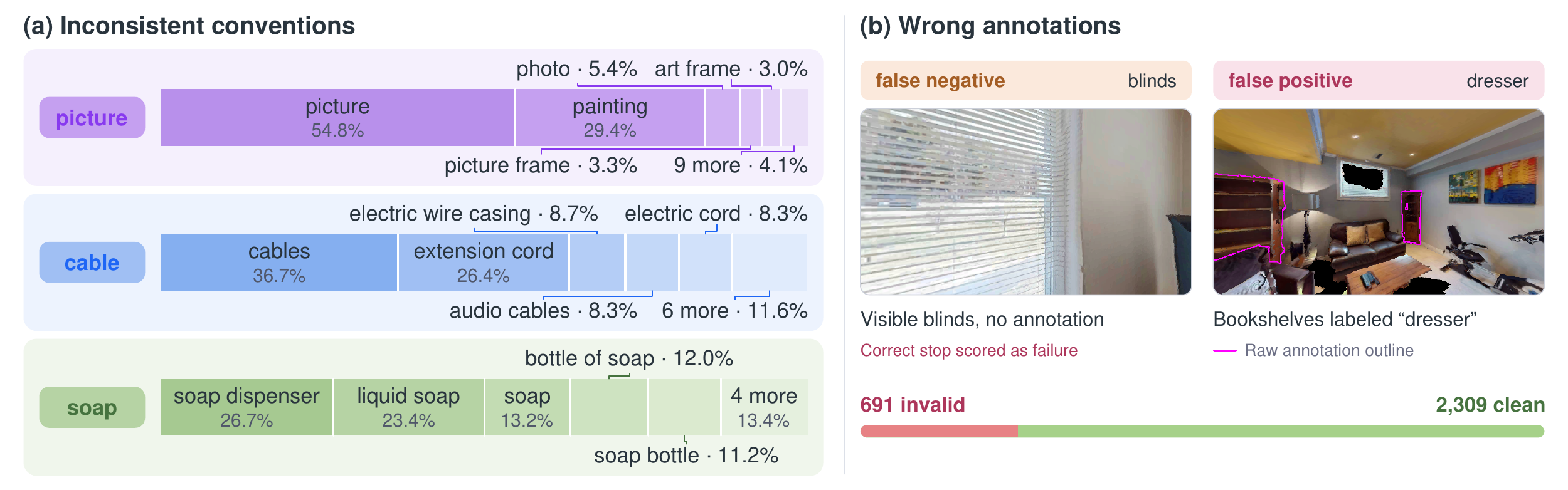}
    \vspace{-6mm}
    \caption{
    \textbf{Two defects of the evaluation.}
    (a) The same object type is annotated under many names, and only the
    exact annotated name earns success credit. (b) The annotations contain
    errors of two kinds, and a manual audit filters out 23\% of the
    val\_unseen episodes.
    }
    \label{fig:vocab}
\end{figure}

\paragraph{Inconsistent conventions.}
Comparing systems relies on evaluating each episode against the object names in the scene annotations. In HM3D, MP3D, and HSSD~\citep{yadav2023hm3dsem, chang2017matterport, khanna2024hssd}, these names are free-form strings with no shared convention. One object type appears under spelling variants and synonyms, and different scenes choose different names for it (\cref{fig:vocab}a). The diversity helps training by exposing the model to a broader vocabulary, but it harms evaluation. Since the benchmark~\citep{yokoyama2024hm3dovon} credits a stop only when the annotation matches the goal word, we find agents scored a failure at perfectly valid goal instances that carry another name, such as a couch annotated as \emph{sofa}, or a picture annotated as \emph{painting}.

\paragraph{Wrong annotations.}
The annotations also contain two kinds of errors (\cref{fig:vocab}b). In a false negative, a visible goal object has no annotation, so a correct stop at it is scored a failure. In a false positive, an annotation gives an object the wrong name, so the agent is sent to an object that is not the goal. Our manual audit filters out about a quarter of HM3D val\_unseen episodes for either error.

\paragraph{Correcting the evaluation.}
\looseness=-1 We repair both defects. For the naming, we merge 2,408 raw labels into 540 object classes and credit a stop at any instance of the goal's class. For the errors, we manually audit the evaluation episodes and drop the affected ones. The corrected criterion counts a stop within 0.25\,m of a viewpoint from which the goal instance fills more than 5\% of the image, so the goal must be clearly visible and some baselines score lower than under the official criterion. The corrected evaluation separates methods more clearly. On HM3D-OVON val\_unseen, it raises LightNav-0 by 13 points, our system by 16, and the same policy with ground-truth grounding by 19 (\cref{tab:main}), because a stronger agent makes more valid stops that the official criterion rejects. Optimizing the official number rewards matching the annotators' naming, not finding the goal.

\section{Experiments}
\label{sec:experiments}

\subsection{Setup}
\label{sec:exp_setup}

\paragraph{Benchmarks.}
We evaluate on open-vocabulary and closed-vocabulary tasks in HM3D, MP3D, and HSSD~\citep{yadav2023hm3dsem,chang2017matterport,khanna2024hssd}. The open-vocabulary tasks are HM3D-OVON~\citep{yokoyama2024hm3dovon}, MP3D-OV, and HSSD-OV. We compose MP3D-OV and HSSD-OV ourselves on the MP3D and HSSD validation scenes, following HM3D-OVON. The closed-vocabulary tasks are ObjectNav~\citep{batra2020objectnav} with 6, 21, and 6 categories. The metric is success rate (SR). We omit SPL here, since the shortest path from the start is no fair reference for a task that requires search, and report it in \cref{app:spl}. Existing benchmarks are scored under two protocols. \emph{Official} uses the benchmark's own criterion. \emph{Clean} uses our corrected evaluation (\cref{sec:vocab}) on the episodes that pass the audit. MP3D-OV and HSSD-OV use only the clean protocol, marked \emph{ours}. The appendix lists episode counts, categories, and simulator settings.

\paragraph{Models and training.}
Our perception model is fine-tuned from Qwen3-VL-4B and the policy from Qwen3-VL-2B~\citep{bai2025qwen3}. Both train on frontier-exploration demonstrations from 145 HM3D-OVON, 61 MP3D, and 123 HSSD training scenes. The corpora contain 3.1M demonstrations (250M frames), before training-index filtering. No evaluation scene is used. The grounding text of every frame comes from the scene annotations. The policy learns actions from it, and the perception model learns to predict it. Training details are in the appendix.

\paragraph{Baselines.}
We compare open-source systems on the same episodes and criteria. Uni-NaVid, LightNav-0, and Nav-R$^2$~\citep{zhang2024uni, wang2026lightnav, xiang2025navr2} are end-to-end VLM policies. DAgRL is the HM3D-OVON baseline~\citep{yokoyama2024hm3dovon}. VLFM~\citep{yokoyama2024vlfm} is a zero-shot modular method. We exclude recent systems with different setups or without released checkpoints~\citep{ren2026astranav, chu2026abot, zhang2026qwen}.

\subsection{Comparison with baselines}
\label{sec:main_results}

\begin{answerbox}{boxlavender}{framelavender}
\textbf{Does explicit goal grounding outperform existing systems?} \textit{Our system leads on all six benchmarks under every protocol, and its lead lies mainly in stopping at the goal.}
\end{answerbox}

\paragraph{Overall comparison.}
\looseness=-1 \Cref{tab:main} evaluates our system and the baselines on all six benchmarks under the same protocols. Our system leads on every benchmark under every reported protocol. The clean one widens the lead, from 6.4 to 8.8 points over LightNav-0 on HM3D-OVON val\_unseen. The clean protocol corrects the annotation defects of \cref{sec:vocab}, so it measures each method more faithfully. Among the two new open-vocabulary benchmarks, our success is higher on HSSD-OV (56.4\%) than on MP3D-OV (49.9\%). Explicit goal grounding pays off even without navigation training. Under the clean protocol, the zero-shot modular VLFM outperforms the fine-tuned end-to-end Uni-NaVid on every task. On HM3D-OVON, our own lead lies mainly in stopping. Once within 0.25\,m of the goal, our system succeeds 90\% of the time on clean val\_unseen and LightNav-0 80\% (\cref{fig:teaser}e).

\begin{table}[t]
\centering
\caption{\looseness=-1 \textbf{Comparison with baselines.} \emph{Official} columns use each benchmark's own success criterion; \emph{clean} columns use our criterion on the audited clean subsets (\cref{sec:exp_setup}). GT grounding is the same policy reading GT grounding text, shown for reference. All numbers are success rates in percent (\%).
}
\vspace{2mm}
\label{tab:main}
\resizebox{\textwidth}{!}{%
\begin{tabular}{l cc cc cc c cc c cc}
\toprule
 & \multicolumn{6}{c}{HM3D} & \multicolumn{3}{c}{MP3D} & \multicolumn{3}{c}{HSSD} \\
\cmidrule(lr){2-7} \cmidrule(lr){8-10} \cmidrule(lr){11-13}
 & \multicolumn{2}{c}{OVON val\_seen} & \multicolumn{2}{c}{OVON val\_unseen} & \multicolumn{2}{c}{6 cat.} & OV & \multicolumn{2}{c}{21 cat.} & OV & \multicolumn{2}{c}{6 cat.} \\
\cmidrule(lr){2-3} \cmidrule(lr){4-5} \cmidrule(lr){6-7} \cmidrule(lr){8-8} \cmidrule(lr){9-10} \cmidrule(lr){11-11} \cmidrule(lr){12-13}
 & official & clean & official & clean & official & clean & ours & official & clean & ours & official & clean \\
\midrule
VLFM~\citep{yokoyama2024vlfm} & 40.0 & 53.6 & 39.0 & 52.8 & 64.7 & 66.5 & 30.8 & 33.6 & 37.8 & \cellcolor{ranktwo}36.3 & 59.0 & \cellcolor{ranktwo}65.1 \\ %
Nav-R$^2$~\citep{xiang2025navr2} & 39.6 & 51.6 & 27.4 & 34.0 & 58.1 & 56.7 & 28.3 & 36.1 & 40.4 & 13.8 & 35.7 & 34.9 \\
DAgRL~\citep{yokoyama2024hm3dovon} & 44.8 & 60.7 & 21.0 & 26.0 & 54.0 & 53.3 & 23.0 & 23.2 & 22.3 & 5.8 & 23.6 & 24.9 \\
Uni-NaVid~\citep{zhang2024uni} & 35.4 & 46.4 & 32.1 & 41.0 & 62.5 & 49.0 & 30.3 & 32.2 & 34.5 & 22.0 & 51.4 & 52.5 \\
LightNav-0~\citep{wang2026lightnav} & \cellcolor{ranktwo}52.0 & \cellcolor{ranktwo}67.4 & \cellcolor{ranktwo}47.5 & \cellcolor{ranktwo}60.8 & \cellcolor{ranktwo}80.6 & \cellcolor{ranktwo}77.7 & \cellcolor{ranktwo}44.3 & \cellcolor{ranktwo}49.0 & \cellcolor{ranktwo}50.8 & 32.8 & \cellcolor{ranktwo}59.9 & 62.9 \\
Ours & \cellcolor{rankone}\textbf{60.4} & \cellcolor{rankone}\textbf{77.6} & \cellcolor{rankone}\textbf{53.9} & \cellcolor{rankone}\textbf{69.6} & \cellcolor{rankone}\textbf{83.0} & \cellcolor{rankone}\textbf{83.6} & \cellcolor{rankone}\textbf{49.9} & \cellcolor{rankone}\textbf{49.9} & \cellcolor{rankone}\textbf{57.9} & \cellcolor{rankone}\textbf{56.4} & \cellcolor{rankone}\textbf{72.8} & \cellcolor{rankone}\textbf{81.2} \\ %
\midrule
\textcolor{gray}{GT grounding} & \textcolor{gray}{79.6} & \textcolor{gray}{92.8} & \textcolor{gray}{72.6} & \textcolor{gray}{91.6} & \textcolor{gray}{86.8} & \textcolor{gray}{90.4} & \textcolor{gray}{74.0} & \textcolor{gray}{66.2} & \textcolor{gray}{73.8} & \textcolor{gray}{77.1} & \textcolor{gray}{74.2} & \textcolor{gray}{90.9} \\ %
\bottomrule
\end{tabular}%
}
\vspace{-4mm}
\end{table}

\begin{figure}[t]
\centering
\includegraphics[width=\textwidth]{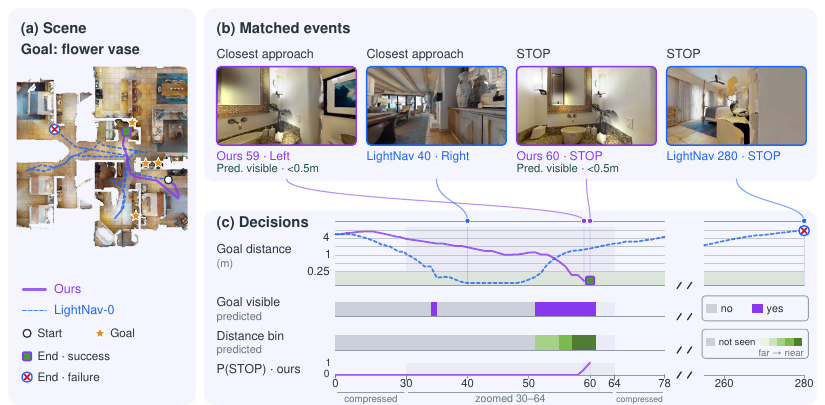}
\vspace{-6mm}
\caption{\textbf{Same episode, different decisions.} (a) Paths of both systems through the scene. (b) Views at each system's closest approach to the goal and at its \texttt{STOP}, labeled with the step and the action taken. Our frames also show the predicted grounding. (c) Goal distance of both systems, with our predicted goal visibility, distance bin, and stop probability. Steps 30 to 64 are enlarged.}
\label{fig:qual_main}
\vspace{-4mm}
\end{figure}

\paragraph{Deciding to stop.}
\Cref{fig:qual_main} follows our system and LightNav-0 through one episode of HM3D-OVON val\_unseen, with the goal ``flower vase''. Both reach the vase, coming within 0.25\,m of it at steps 59 and 40. From step 57, our perception reports the vase in view within 0.5\,m. The stop probability then rises, and the policy stops at step 60. LightNav-0 has the vase in view at step 40 but turns away, and stops 7.4\,m from it at step 280. \Cref{fig:teaser}e shows that the end-to-end baselines often fail this way, like our end-to-end policy in \cref{sec:failure}.

\subsection{Analysis}
\label{sec:analysis}

\begin{answerbox}{boxgreen}{framegreen}
\textbf{Where is the remaining headroom?} \textit{Every design choice contributes, disjoint grounding the most. With ground-truth grounding, the same policy nearly matches the expert, so the remaining headroom lies in perception rather than policy learning.}
\end{answerbox}

\begin{wraptable}[11]{r}{0.4\textwidth}
\vspace{-4mm}
\centering
\small
\caption{\textbf{Ablation study.} Success of the design choices on the clean OVON splits. The last row is our complete design and has the best performance. }
\label{tab:ablation}
\vspace{1mm}
\setlength{\tabcolsep}{3pt}
\resizebox{\linewidth}{!}{%
\begin{tabular}{ccc|cc}
\toprule
Disjoint & \begin{tabular}[c]{@{}c@{}}Semantic\\grounding\end{tabular} & \begin{tabular}[c]{@{}c@{}}Distance\\grounding\end{tabular} & Seen & Unseen \\
\midrule
 &  &  & 72.7 & 63.6 \\
 & $\checkmark$ & $\checkmark$ & 64.4 & 50.7 \\
$\checkmark$ & $\checkmark$ &  & 76.5 & 68.7 \\
$\checkmark$ &  & $\checkmark$ & 73.4 & 60.1 \\ %
\rowcolor{rankone}$\checkmark$ & $\checkmark$ & $\checkmark$ & 77.6 & 69.6 \\
\bottomrule
\end{tabular}}
\end{wraptable}

\paragraph{Ablation study.}
We ablate our design choices in \cref{tab:ablation}. Disjoint grounding matters most. When one model writes the grounding text and then acts, success on val\_unseen falls to 50.7\%, even below the policy without any grounding. A separate perception model writing the same text raises it to 69.6\%. Grounding text thus helps only when a separate model writes it. The semantic field alone recovers most of this gain (68.7\%), and the distance field further improves success. Each field tells the policy whether it has found the goal, which the decision to stop depends on. Both fields together do best.

\begin{figure}[t]
\centering
\includegraphics[width=0.9\textwidth]{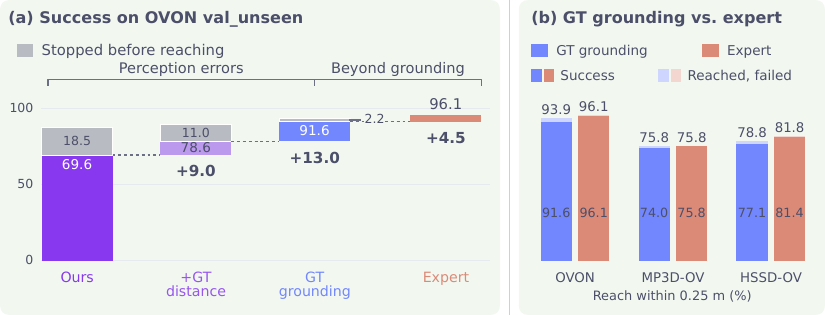}
\vspace{-3mm}
\caption{\textbf{From our system to the expert.} (a) Success on HM3D-OVON val\_unseen as predicted grounding fields are replaced with ground truth. Gray marks episodes where the policy stops before ever coming within 0.25\,m of the goal. (b) The policy with ground-truth grounding against the expert. Bar height is reach within 0.25\,m, and the solid part is success. Clean subsets throughout.}
\label{fig:ladder}
\vspace{-4mm}
\end{figure}

\paragraph{Perception as the bottleneck.}
Most of the remaining failures in our system come from perception. With ground-truth grounding, the same policy's success on clean HM3D-OVON val\_unseen rises by 22.0 points (\cref{tab:main}). To find which perception errors cost these points, we replace the predicted grounding with ground truth in two steps (\cref{fig:ladder}a). The first step corrects only the goal distance when both the predicted and ground-truth inventories contain the goal, raising success by 9.0 points to 78.6\%. The second step also corrects the category list, adding 13.0 points to reach 91.6\%. The remaining gap to the expert is 4.5 points. Better goal recognition and better distance estimation can both carry over to navigation without retraining the policy.

\paragraph{Search with correct grounding.}
We next ask whether the policy itself still limits the system once grounding is correct. We compare the policy given ground-truth grounding with the frontier-exploration expert that generated its demonstrations, on the same episodes (\cref{fig:ladder}b). On all three benchmarks, the policy comes within 0.25\,m of the goal about as often as the expert, and most of its remaining gap arises after arrival, when it fails to stop. With correct grounding, the policy searches nearly as well as its teacher in new scenes of all three collections.

\paragraph{Evaluation ceiling.}
We last ask how much headroom remains above the policy with ground-truth grounding. We evaluate the same expert on HM3D-OVON val\_unseen. The clean protocol credits the expert in 96\% of the episodes. Nearly all of its failures come from running out of frontiers or steps before it finds the goal. The official protocol credits it in 72\%, about the same as the policy with ground-truth grounding. The difference consists of expert stops at objects that our vocabulary accepts as the goal but the official protocol rejects (\cref{fig:vocab}). Under the official protocol, most of the remaining headroom therefore reflects the evaluation rather than navigation.

\subsection{Perception errors}
\label{sec:remaining}

\begin{answerbox}{boxyellow}{frameyellow}
\textbf{Which perception errors cause the remaining failures?} \textit{Wrong goal visibility and distance cause most failures, and per-category success rises with how often perception reports the goal.}
\end{answerbox}

\begin{figure}[t]
    \centering
    \includegraphics[width=\textwidth]{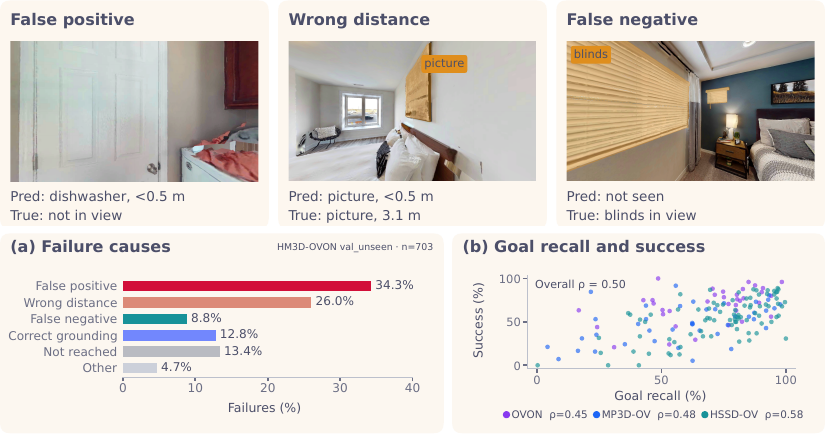}
    \vspace{-6mm}
    \caption{
    \textbf{Perception errors of our system.}
    Top, examples of each error type. (a) Final-system failures on OVON val\_unseen by cause. (b) Final-system success per goal category against how often the perception model reports the goal in view.
    }
    \label{fig:remaining}
\end{figure}

\paragraph{Error types.}
\Cref{sec:analysis} places most of the remaining gap in perception. We now ask which perception errors cause these failures. For each failed episode on HM3D-OVON val\_unseen, we check what the perception model reported when the outcome was decided, at the \texttt{STOP} step or at a timeout near the goal. As shown in \cref{fig:remaining}a, three errors account for 69\% of the failures. In a false positive, the perception model reports the goal although it is not in view. In a wrong distance, it underestimates the distance bin of a visible goal. In a false negative, it does not report the goal when it is clearly in view (\cref{fig:remaining}, top). Wrong goal visibility causes 43\% of the failures, and wrong goal distance causes 26\%. Both parts of goal grounding still limit the system.

\paragraph{Correlation across categories.}
We last ask whether perception also explains why success differs across goal categories. For each category, we compare its success rate with how often the perception model reports the goal when it is in view. On all three benchmarks, success rises with this rate (\cref{fig:remaining}b). The least reported goals are small or thin objects, such as hangers, bottles, and books, while large furniture such as beds and couches is reported reliably. With the ground-truth substitution in \cref{sec:analysis}, this correlation indicates that higher success needs better perception.

\section{Discussion and Conclusion}
\label{sec:conclusion}

\paragraph{Broader impact.}
Our work offers three things. First, it identifies goal grounding as the capability that limits VLM navigation policies, and it shows that grounding can be supervised from scene annotations and improved apart from the policy. Second, it shows that the official evaluations reject valid stops, and more of them for stronger agents (\cref{sec:vocab}). Comparisons between methods therefore need a clean evaluation, and we release the corrected one. Third, we release two new open-vocabulary benchmarks, MP3D-OV and HSSD-OV, which extend evaluation beyond HM3D.

\paragraph{Limitations.}
Our work has two main limitations. First, the grounding supervision comes from the semantic and geometric annotations of 3D scenes, so training data is limited to annotated scene collections. Learning grounding from real videos without such annotations would open much larger data sources. Second, we evaluate only in simulation, on object-goal navigation in static indoor scenes. We expect our gains to carry over to real robots, where VLM policies trained mainly on the same simulated scenes, such as LightNav-0, already run without adaptation~\citep{wang2026lightnav}.

\paragraph{Conclusion.}
Recent VLM navigation policies have grown through more demonstrations and better history modeling, yet generalization to unseen goals still lags. We traced this gap to goal grounding. The agent learns to search from demonstrations but often does not stop at the goal, because it cannot tell whether the goal is in view and how far away it is. Explicit grounding splits navigation into a policy that nearly matches its teacher once grounding is correct and a perception model that accounts for most of the remaining failures. Improvements to that perception model carry over to navigation without retraining the policy. Our two-model pipeline is one way to supply goal grounding. A single model with a grounding objective or an agentic system that reasons about the goal could also supply it, and our clean benchmarks let future work compare these routes. We expect object-goal navigation to advance with goal grounding.

\bibliography{iclr2027_conference}
\bibliographystyle{iclr2027_conference}

\appendix
\crefalias{section}{appendix}
\crefalias{subsection}{subappendix}
\clearpage
\setcounter{topnumber}{3} %
\section{Comparison on Perception Models}
\label{sec:vlm_perception}

\begin{answerbox}{boxflamingo}{frameflamingo}
\textbf{Does a stronger perception model raise success with the same policy?} \textit{Yes. A stronger VLM misses fewer goals than our perception model and raises success with the same policy.}
\end{answerbox}

\begin{figure}[ht]
\centering
\includegraphics[width=\textwidth]{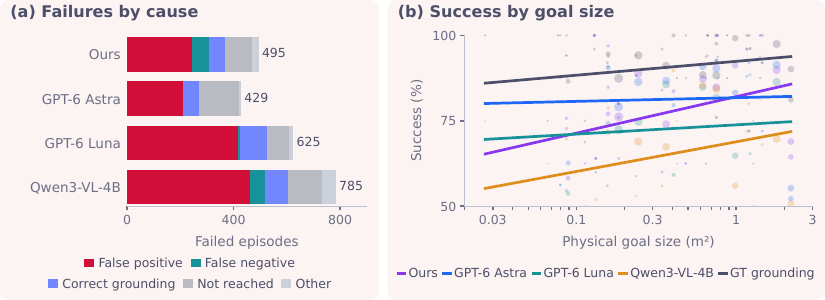}
\vspace{-4mm}
\caption{\textbf{Failures and goal size with VLM grounding.} (a) Failed episodes by perception error, as in \cref{fig:remaining}a. (b) Success per goal category against its physical size, the median area of its training instances, with an episode-weighted linear fit for each source.}
\label{fig:vlm_perception}
\vspace{-6mm}
\end{figure}

\begin{table}[ht]
\centering
\small
\setlength{\tabcolsep}{5pt}
\caption{\textbf{Comparison on perception models.} Goal recall and object F1 on OVON val\_unseen frames, and success of the same policy on clean val\_unseen with GT goal distance as in the first step of \cref{fig:ladder}a. The VLMs are not trained on our task. Among them, success rises with goal recall. $^*$Amazon Bedrock API, with the second GPT-6 Astra value on its ultrafast API.}
\label{tab:vlm}
\begin{tabular}{@{}lcccc}
\toprule
Perception model & Goal recall & Object F1 & Success & Step time (s) \\
\midrule
Qwen3-VL-4B & \cellcolor{frameflamingo!26}83.2 & \cellcolor{frameflamingo!6}48.6 & \cellcolor{frameflamingo!6}66.0 & \cellcolor{frameflamingo!46}1.0 \\
Joint model & \cellcolor{frameflamingo!6}70.0 & \cellcolor{frameflamingo!37}66.4 & -- & -- \\
GPT-6 Luna & \cellcolor{frameflamingo!36}89.7 & \cellcolor{frameflamingo!22}57.8 & \cellcolor{frameflamingo!24}72.9 & \cellcolor{frameflamingo!36}6.4$^*$ \\
GPT-6 Astra & \cellcolor{frameflamingo!46}95.9 & \cellcolor{frameflamingo!24}58.6 & \cellcolor{frameflamingo!46}81.4 & \cellcolor{frameflamingo!6}16.6 / 9.3$^*$ \\
Ours & \cellcolor{frameflamingo!18}77.9 & \cellcolor{frameflamingo!46}71.3 & \cellcolor{frameflamingo!39}78.6 & \cellcolor{frameflamingo!46}1.0 \\
\midrule
Ground-truth grounding & -- & -- & 91.1 & -- \\
\bottomrule
\end{tabular}
\end{table}

\paragraph{Stronger perception.}
\Cref{sec:remaining} attributes most remaining failures to perception errors. We now ask whether a stronger perception model raises success with the same policy. We replace our perception model with three VLMs that are not trained on our task (\cref{tab:vlm}). GPT-6 Astra outperforms our perception model, so the policy turns better perception into higher success without retraining. Astra is about nine times slower per step even on its Ultrafast tier, however, so it is not a practical replacement. Training on our task still matters at a fixed model size, as our perception model far outperforms the untrained Qwen3-VL-4B it is fine-tuned from. Even models with strongest grounding stay well below GT grounding, so better perception still has room to raise success.

\paragraph{Recall and false positives.}
We next ask what makes one grounding source better than another. Among the VLMs, success follows goal recall. Recall also explains why GPT-6 Astra leads our perception model. Our perception model often misses a visible goal even after the agent reaches it, a false negative that GPT-6 Astra almost never makes (\cref{fig:vlm_perception}a). GPT-6 Luna and Qwen3-VL-4B also recall more goals than our perception model, yet they succeed less. Their lower object F1 shows that they also report objects the annotations do not contain, and the resulting false positives outweigh their higher recall. Higher recall thus helps only when it does not bring more false positives. The joint model recalls fewer goals than our perception model, consistent with its lower success in \cref{tab:ablation}.

\paragraph{Goal size.}
We last ask which goals remain hard (\cref{fig:vlm_perception}b). Small goals lower success most with our perception model, while GPT-6 Astra succeeds almost equally across sizes. The lead of Astra thus comes mostly from small goals. With ground-truth grounding, success depends far less on size than with our perception model. Small goals are therefore hard to perceive rather than to reach.

\clearpage
\section{Path Efficiency}
\label{app:spl}

Prior work also reports success weighted by path length (SPL)~\citep{anderson2018evaluation}. For episode $i$ with success $S_i$, agent path length $p_i$, and reference length $\ell_i$,
\[
\mathrm{SPL} = \frac{1}{N}\sum_{i=1}^{N} S_i\,\frac{\ell_i}{\max(\ell_i,\,p_i)}.
\]
SPL takes $\ell_i$ as the shortest-path distance from the start to the goal. This reference assumes that the agent already knows where the goal is, which no agent does on a task that requires search. We therefore also report SPL$_\mathrm{FE}$, which replaces $\ell_i$ with the length of the path that the frontier-exploration expert travels on the same episode. Our system leads every baseline in all three metrics on all three benchmarks (\cref{tab:app_spl}).

\begin{table}[t]
\centering
\footnotesize
\setlength{\tabcolsep}{3.5pt}
\caption[Path efficiency]{Path efficiency on the clean open-vocabulary benchmarks.}
\label{tab:app_spl}
\begin{tabular}{@{}l ccc ccc ccc@{}}
\toprule
 & \multicolumn{3}{c}{OVON val\_unseen} & \multicolumn{3}{c}{MP3D-OV} & \multicolumn{3}{c}{HSSD-OV} \\
\cmidrule(lr){2-4} \cmidrule(lr){5-7} \cmidrule(lr){8-10}
 & SR & SPL & SPL$_\mathrm{FE}$ & SR & SPL & SPL$_\mathrm{FE}$ & SR & SPL & SPL$_\mathrm{FE}$ \\
\midrule
VLFM~\citep{yokoyama2024vlfm} & 52.8 & \cellcolor{ranktwo}30.2 & \cellcolor{ranktwo}39.8 & 30.8 & 16.7 & 27.1 & \cellcolor{ranktwo}36.3 & \cellcolor{ranktwo}18.5 & \cellcolor{ranktwo}29.0 \\
Nav-R$^2$~\citep{xiang2025navr2} & 34.0 & 14.4 & 19.9 & 28.3 & 11.0 & 19.6 & 13.8 & 6.2 & 9.8 \\
DAgRL~\citep{yokoyama2024hm3dovon} & 26.0 & 10.7 & 15.7 & 23.0 & 9.6 & 17.8 & 5.8 & 1.7 & 3.5 \\
Uni-NaVid~\citep{zhang2024uni} & 41.0 & 16.7 & 23.0 & 30.3 & 12.7 & 22.2 & 22.0 & 10.1 & 16.9 \\
LightNav-0~\citep{wang2026lightnav} & \cellcolor{ranktwo}60.8 & 28.0 & 38.0 & \cellcolor{ranktwo}44.3 & \cellcolor{ranktwo}21.4 & \cellcolor{ranktwo}34.9 & 32.8 & 16.9 & 25.7 \\
Ours & \cellcolor{rankone}\textbf{69.6} & \cellcolor{rankone}\textbf{36.5} & \cellcolor{rankone}\textbf{50.7} & \cellcolor{rankone}\textbf{49.9} & \cellcolor{rankone}\textbf{24.0} & \cellcolor{rankone}\textbf{40.2} & \cellcolor{rankone}\textbf{56.4} & \cellcolor{rankone}\textbf{27.7} & \cellcolor{rankone}\textbf{44.9} \\
\midrule
\textcolor{gray}{GT grounding} & \textcolor{gray}{91.6} & \textcolor{gray}{48.8} & \textcolor{gray}{68.1} & \textcolor{gray}{74.0} & \textcolor{gray}{36.9} & \textcolor{gray}{60.1} & \textcolor{gray}{77.1} & \textcolor{gray}{36.3} & \textcolor{gray}{59.6} \\
\bottomrule
\end{tabular}
\end{table}

\section{Benchmarks and Simulation}
\label{app:benchmarks}

\Cref{tab:app_benchmarks} lists the evaluation sets, all on scenes held out from training.
HM3D-OVON and the closed-vocabulary tasks use official validation episodes.
MP3D-OV and HSSD-OV use episodes generated on publisher validation scenes, with category goals from our vocabulary.
Every clean subset also excludes episodes whose destination set under our criterion is empty or has no finite path from the start on the protocol navmesh, independently of outcomes.
Sixteen baseline successes fall on these excluded episodes, six for LightNav-0 on HM3D ObjectNav and ten for Nav-R$^2$ on HSSD ObjectNav, where the baselines reached the goals with their own embodiments.

\begin{table}[t]
\centering
\small
\setlength{\tabcolsep}{4pt}
\caption{Evaluation sets. Episodes, scenes, goal names, and goal groups describe the raw sets before exclusions. Goal groups count equivalence groups under our vocabulary. Clean counts give the adopted evaluation subsets.}
\label{tab:app_benchmarks}
\begin{tabular}{@{}lrrrrr@{}}
\toprule
Evaluation set & Episodes & Scenes & Goal names & Goal groups & Clean \\
\midrule
HM3D-OVON val\_seen & 3,000 & 36 & 79 & 66 & 1,937 \\
HM3D-OVON val\_unseen & 3,000 & 36 & 49 & 40 & 2,309 \\
HM3D ObjectNav six-class & 1,000 & 36 & 6 & n/a & 905 \\
MP3D-OV & 3,000 & 11 & 62 & 59 & 1,193 \\
MP3D ObjectNav 21-class & 2,195 & 11 & 21 & n/a & 1,078 \\
HSSD-OV & 3,000 & 43 & 110 & 104 & 1,751 \\
HSSD ObjectNav six-class & 1,248 & 42 & 6 & n/a & 909 \\
\bottomrule
\end{tabular}
\end{table}

Generated goals must have annotated instances and valid starts in multiple validation scenes.
We sample navigable starts and uniform yaw angles, with the nearest accepted target reachable at a geodesic distance between 1 and 30\,m.

\paragraph{Simulation.}
Our system uses the Stretch embodiment in \cref{tab:app_simulation}.
The action set is \texttt{Forward}, \texttt{Left}, \texttt{Right}, and \texttt{Stop}.
Episodes that exhaust the action budget without stopping fail.

\begin{table}[t]
\centering
\small
\caption{Simulation settings for our system. Image dimensions are width by height.}
\label{tab:app_simulation}
\begin{tabular}{@{}ll@{}}
\toprule
Parameter & Value \\
\midrule
Body height / radius & 1.41\,m / 0.17\,m \\
Camera height & 1.31\,m \\
Rendered RGB / horizontal field of view & $640\times360$ / $120^\circ$ \\
Forward step / turn angle & 0.25\,m / $30^\circ$ \\
Action budget / sliding & 500 / enabled \\
\bottomrule
\end{tabular}
\end{table}

\section{Qualitative Comparisons}
\label{app:qual}

\begin{figure}[p]
\centering
\includegraphics[width=0.93\textwidth]{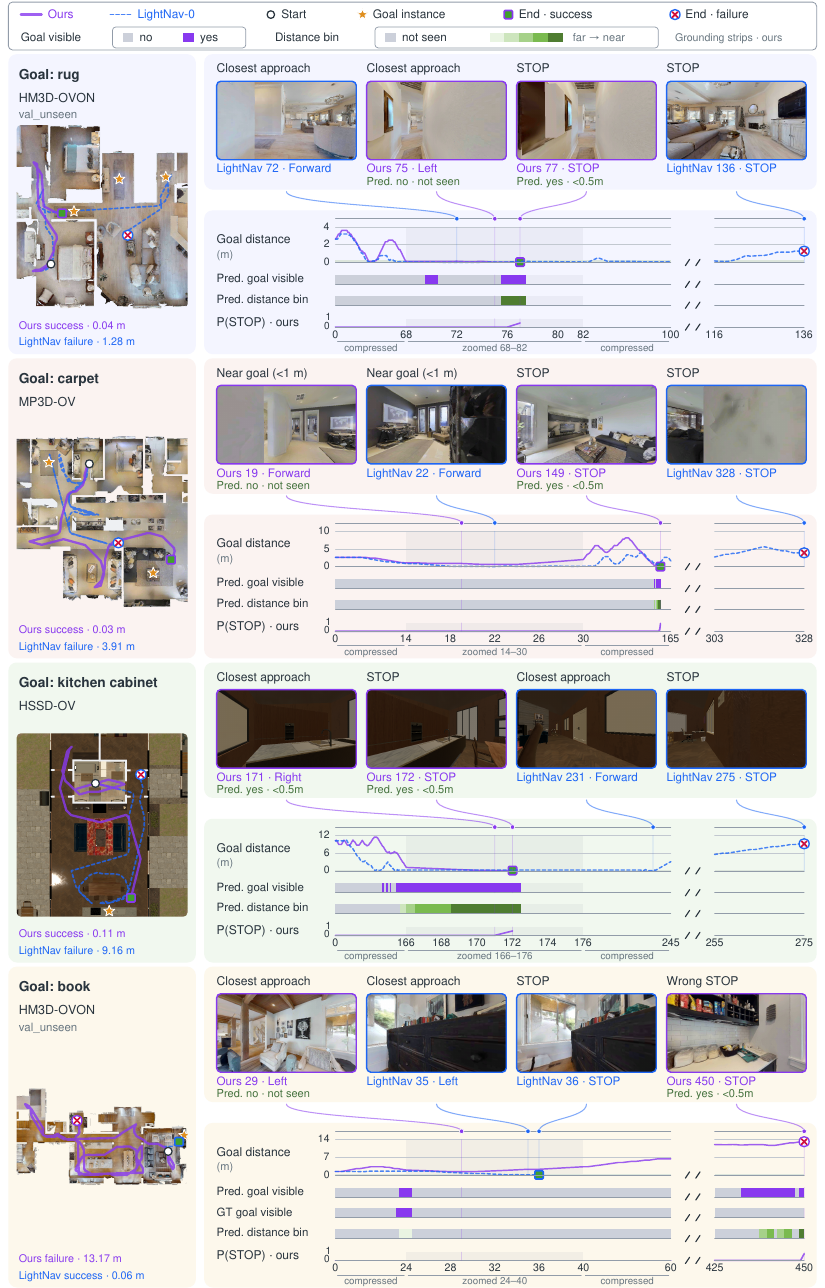}
\caption{Same-episode comparisons of our system and LightNav-0. Each row follows the layout of \cref{fig:qual_main}. The last row is a failure of ours and adds the ground-truth goal visibility.}
\label{fig:qual_appendix}
\end{figure}

\Cref{fig:qual_appendix} shows four more episodes on which our system and LightNav-0 diverge.
We selected one episode per open-vocabulary benchmark and one failure of ours from candidates ranked by their distance curves.
In the rug episode (HM3D-OVON val\_unseen), LightNav-0 comes within 0.01\,m of the rug, continues forward, and stops 1.3\,m away at step 136. Our policy stops at step 77, 0.04\,m from it.
In the carpet episode (MP3D-OV), both systems pass within 0.25\,m of the carpet early. Our system returns and stops at step 149, and LightNav-0 stops 3.9\,m away at step 328.
In the kitchen cabinet episode (HSSD-OV), LightNav-0 comes within 0.02\,m, continues, and stops 9.2\,m away at step 275. Our policy stops at step 172, 0.11\,m from it.
In the book episode (HM3D-OVON val\_unseen), our perception reports the book in view within 0.5\,m at step 450 while it is not in view. Our policy stops there, 13.2\,m from the goal. LightNav-0 stops 0.06\,m from it at step 36.

\section{Corrected Evaluation}
\label{app:eval}

Our vocabulary maps scene labels to shared object classes and groups equivalent goal names.
Our success criterion requires \texttt{Stop} below 0.25\,m geodesic distance to any accepted instance viewpoint, measured on the Stretch navigation mesh.
An instance viewpoint is one from which the goal instance fills more than 5\% of the image, summed over three camera pitches.
Official success uses the benchmark's own goal viewpoints.
MP3D-OV and HSSD-OV use only our criterion.

We manually audit the annotations by comparing rendered frames with their semantic annotations.
An annotation false negative is a visible goal object with no annotation; an annotation false positive is an object annotated as a goal that it is not.
We flag scene/category pairs with repeated evidence of either error and exclude episodes whose goal is flagged in their scene.
MP3D-OV and HSSD-OV also exclude broad goal names that do not specify an object type.
The clean protocol combines these exclusions with our success criterion; retained counts appear in \cref{tab:app_benchmarks}.

\section{Grounding Text and Expert Demonstrations}
\label{app:grounding}

\paragraph{Grounding text.}
For ObjectNav, the policy reads the fields in \cref{tab:app_text} at each observation.
The inventory contains at most 16 categories ordered by decreasing image area.
Categories pass area thresholds derived from training-instance geometry, with the goal retained whenever it passes its threshold.
Goal visibility is inventory membership, and distance is supplied only when the goal is visible in the current image.

\begin{table}[t]
\centering
\small
\caption{Grounding text read by the policy. The action prompt follows these fields.}
\label{tab:app_text}
\begin{tabular}{@{}lp{0.71\linewidth}@{}}
\toprule
Field & Format \\
\midrule
\texttt{Objects:} & Category names separated by \texttt{ | } \\
\texttt{Goal visible:} & \texttt{true} or \texttt{false} \\
\texttt{Distance:} & \texttt{<0.5m}, \texttt{0.5-1m}, \texttt{1-2m}, \texttt{2-4m}, \texttt{>4m}, or \texttt{not seen} \\
\bottomrule
\end{tabular}
\end{table}

Perception predicts the inventory and distance bin used to construct these inputs.

\paragraph{Frontier-exploration expert.}
The expert explores until it can approach an accepted goal instance.
Its destination set contains viewpoints adjacent to accepted goal instances, and the distance label is the geodesic distance to the nearest destination.
The expert stops within 0.15\,m of its destination after facing the object.

\paragraph{Training corpora.}
The policy's corpora provide 3,138,329 demonstrations and 252,526,115 frames from 145 HM3D, 61 MP3D, and 123 HSSD training scenes.
The perception model uses frontier-exploration demonstrations, totaling 3,140,902 demonstrations and 250,013,894 frames from the same 145 HM3D, 61 MP3D, and 123 HSSD training scenes.
These totals precede training-index filtering.

\section{Training Details}
\label{app:training}

\Cref{tab:app_training} summarizes training from the Qwen3-VL initializations~\citep{bai2025qwen3}.

\begin{table}[t]
\centering
\small
\setlength{\tabcolsep}{4pt}
\caption{Training hyperparameters. Frame budgets are global caps per update. FSDP is fully sharded data parallelism. Training time sums the running time of all job attempts, including setup and checkpointing. The policy run resumed twice after preemption.}
\label{tab:app_training}
\begin{tabular}{@{}p{0.40\linewidth}p{0.26\linewidth}p{0.26\linewidth}@{}}
\toprule
Parameter & Perception & Policy \\
\midrule
Initialization & Qwen3-VL-4B-Instruct & Qwen3-VL-2B-Instruct \\
Model RGB input & $640\times352$ & $320\times160$ \\
Maximum history frames & 128 & 500 \\
Updates & 20,000 & 40,000 \\
Frames per update, cap & 8,192 & 8,192 \\
Peak learning rate & $10^{-5}$ & $10^{-5}$ \\
Learning-rate schedule & Cosine & Cosine \\
\midrule
GPUs & 64 B200 & 16 B200 \\
Parallelism & FSDP & FSDP \\
Precision & BF16, FP32 master weights & BF16, FP32 master weights \\
Training time & 11.6\,h & 17.9\,h \\
GPU-hours & 743 & 286 \\
\bottomrule
\end{tabular}
\end{table}

Perception reads RGB and expert-action history during training.
It learns to predict grounding text, while the policy learns expert actions from RGB, ground-truth grounding text, and prior actions.
The models are trained separately.

\section{Failure Analysis Protocol}
\label{app:failure}

\Cref{fig:failure} compares the same Qwen3-VL-2B policy trained twice on HM3D-OVON expert demonstrations for 20,000 updates.
Both runs use $320\times160$ model inputs and up to 500 frames of history.
One reads RGB and prior actions.
The other also reads ground-truth grounding text with distance-bin edges at 0.15, 0.25, 0.5, 1, 2, and 4\,m.
Both cover all 3,000 HM3D-OVON val\_unseen episodes.

\paragraph{Reaching a goal.}
An episode reaches the goal vicinity when any saved trajectory position has geodesic distance below 0.25\,m to an accepted instance viewpoint.
Distances are measured on the Stretch navigation mesh over all equivalent goal instances under our vocabulary.
The same target set determines success; the grounding distance instead uses the adjacent viewpoints in \cref{app:grounding}.

\paragraph{Outcomes.}
Success requires \texttt{Stop} below 0.25\,m geodesic distance to the accepted viewpoint set.
A wrong stop is a stop that fails this criterion.
A timeout executes 500 actions without stopping.
\Cref{tab:app_failure} partitions these outcomes by reach, with all percentages using 3,000 episodes as the denominator.
Failure after reaching includes both wrong stops and timeouts in the reached branch.

\begin{table}[t]
\centering
\small
\caption{Counts behind \Cref{fig:failure}. The ground-truth-grounding policy is the same single-source policy described above.}
\label{tab:app_failure}
\begin{tabular}{@{}lrr@{}}
\toprule
Branch or outcome & RGB-only & Ground-truth grounding \\
\midrule
All episodes & 3,000 & 3,000 \\
Reached vicinity & 2,288 & 2,818 \\
\quad Success & 1,569 & 2,679 \\
\quad Wrong stop & 183 & 0 \\
\quad Timeout & 536 & 139 \\
Never reached vicinity & 712 & 182 \\
\quad Wrong stop & 507 & 0 \\
\quad Timeout & 205 & 182 \\
\midrule
Reached vicinity / all episodes & 76.27\% & 93.93\% \\
Success / all episodes & 52.30\% & 89.30\% \\
Failed after reaching / all episodes & 23.97\% & 4.63\% \\
\bottomrule
\end{tabular}
\end{table}

\begin{figure}[t]
\centering
\includegraphics[width=\textwidth]{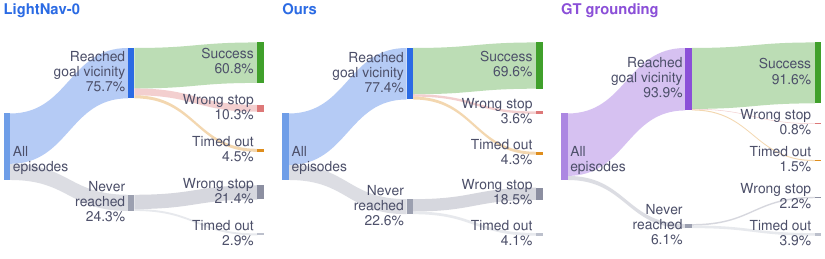}
\caption[Failure flows of the final systems]{Failure flows on clean HM3D-OVON val\_unseen, as in~\cref{fig:failure}. Percentages use the 2,309 clean episodes.}
\label{fig:app_sankey}
\end{figure}

\paragraph{Final systems.}
\Cref{fig:app_sankey} applies the same partition to LightNav-0, our system, and the same policy with ground-truth grounding on the 2,309 clean HM3D-OVON val\_unseen episodes. Seven LightNav-0 episodes that ended on a simulator crash count as timeouts. LightNav-0 reaches the goal vicinity about as often as our system (75.7\% and 77.4\%), but it stops in the wrong place after reaching nearly three times as often (10.3\% and 3.6\%). With ground-truth grounding, the policy reaches the vicinity in 93.9\% of the episodes and fails after reaching in 2.3\%.

\paragraph{Grounding ladder and expert reference.}
\Cref{fig:ladder} uses the same 4B perception model and 2B policy as \cref{tab:main}.
The middle rung keeps predicted categories and substitutes ground-truth distance only when both inventories contain the goal.
The full-grounding condition gives the same policy the annotated inventory and distance.
The expert is the controller that generated our demonstrations.
Under the official protocol, it succeeds in 2,152 of all 3,000 HM3D-OVON val\_unseen episodes (72\%).
\Cref{tab:app_ladder} gives the clean-subset counts, with reach counting positions below 0.25\,m from a goal viewpoint.
Episodes without usable goal viewpoints are excluded from the clean subsets.

\begin{table}[t]
\centering
\small
\caption{Clean-subset counts for \cref{fig:ladder}. Pred.\ uses predicted grounding, GT uses ground-truth grounding, and Exp.\ denotes the frontier-exploration expert.}
\label{tab:app_ladder}
\resizebox{\textwidth}{!}{%
\begin{tabular}{@{}l *{9}{>{\raggedleft\arraybackslash}p{2.8em}}@{}}
\toprule
 & \multicolumn{3}{c}{\begin{tabular}[b]{@{}c@{}}HM3D-OVON val\_unseen\\2,309\end{tabular}} & \multicolumn{3}{c}{\begin{tabular}[b]{@{}c@{}}MP3D-OV\\1,193\end{tabular}} & \multicolumn{3}{c}{\begin{tabular}[b]{@{}c@{}}HSSD-OV\\1,751\end{tabular}} \\
\cmidrule(lr){2-4} \cmidrule(lr){5-7} \cmidrule(l){8-10}
Outcome & Pred. & GT & Exp. & Pred. & GT & Exp. & Pred. & GT & Exp. \\
\midrule
Success & 1,606 & 2,115 & 2,218 & 595 & 883 & 904 & 988 & 1,350 & 1,426 \\
Reached, wrong stop & 82 & 18 & 0 & 46 & 5 & 0 & 31 & 7 & 1 \\
Reached, timeout & 100 & 34 & 1 & 45 & 16 & 0 & 41 & 23 & 6 \\
Never reached, stop & 427 & 51 & 63 & 284 & 9 & 98 & 408 & 25 & 104 \\
Never reached, timeout & 94 & 91 & 27 & 223 & 280 & 191 & 283 & 346 & 214 \\
No usable goal viewpoint & & & 0 & & & 0 & & & 0 \\
\midrule
Reached (\%) & 77.4 & 93.9 & 96.1 & 57.5 & 75.8 & 75.8 & 60.5 & 78.8 & 81.8 \\
Success (\%) & 69.6 & 91.6 & 96.1 & 49.9 & 74.0 & 75.8 & 56.4 & 77.1 & 81.4 \\
\bottomrule
\end{tabular}%
}
\end{table}

\paragraph{Failure attribution.}
\Cref{fig:remaining} assigns each of the 703 failures on the 2,309 clean HM3D-OVON val\_unseen episodes one cause.
We inspect the final pre-action frame at a stop, or the interval from first reach through the end of a reached timeout.
Reach uses the ladder's rule of a distance below 0.25\,m.
A false positive is a failed stop that reports a goal not visible under the annotation-derived visibility rule.
Wrong distance is a failed stop with a visible, reported goal and an underestimated distance bin.
A false negative is a reached timeout with a visible goal but no simultaneous true-positive report in the inspected interval.
Correct grounding includes failed stops with correct visibility and distance bins, reached timeouts with a true-positive goal report and predicted distance below 0.5\,m, and never-reached stops with the goal neither visible nor reported.
Not reached contains never-reached timeouts.
Other contains unforeseen patterns and unresolved cases, including reached timeouts with no visible goal.
The six counts, in this order, are 241, 183, 62, 90, 94, and 33.
For the category-level analysis, we retain categories with at least ten clean episodes and compute report recall over all frames where the annotated goal is visible.

\end{document}